\documentclass{sig-alternate-2013}

\newfont{\mycrnotice}{ptmr8t at 7pt}
\newfont{\myconfname}{ptmri8t at 7pt}
\let\crnotice\mycrnotice%
\let\confname\myconfname%

\permission{Permission to make digital or hard copies of all or part of this work for personal or classroom use is granted without fee provided that copies are not made or distributed for profit or commercial advantage and that copies bear this notice and the full citation on the first page. Copyrights for components of this work owned by others than ACM must be honored. Abstracting with credit is permitted. To copy otherwise, or republish, to post on servers or to redistribute to lists, requires prior specific permission and/or a fee. Request permissions from permissions@acm.org.}
\conferenceinfo{ICMR'15,}{June 23--26, 2015, Shanghai, China.\\
{\mycrnotice{Copyright is held by the owner/author(s). Publication rights licensed to ACM.}}}
\copyrightetc{ACM \the\acmcopyr}
\crdata{978-1-4503-3274-3/15/06\ ...\$15.00.\\
DOI: http://dx.doi.org/10.1145/2671188.2749408
}

\begin{document}
%
% --- Author Metadata here ---
%\conferenceinfo{ICMR}{2015, Shanghai, China}
%\CopyrightYear{2015} % Allows default copyright year (20XX) to be over-ridden - IF NEED BE.
%\crdata{978-1-4503-3274-3/15/06}  % Allows default copyright data (0-89791-88-6/97/05) to be over-ridden - IF NEED BE.
% --- End of Author Metadata ---

\title{Multi-view Face Detection Using Deep Convolutional Neural Networks}


\numberofauthors{3} %  in this sample file, there are a *total*
\author{
% 1st. author
\alignauthor
Sachin Sudhakar Farfade\\
       \affaddr{Yahoo}\\
       \email{fsachin@yahoo-inc.com}\\ 
% 2nd. author
\alignauthor
Mohammad Saberian\\
       \affaddr{Yahoo}\\
       \email{saberian@yahoo-inc.com}\\ 
% 3rd. author
\alignauthor 
Li-Jia Li\\
       \affaddr{Yahoo}\\
       \email{lijiali.vision@gmail.com}\\ 
}
% There's nothing stopping you putting the seventh, eighth, etc.
% author on the opening page (as the 'third row') but we ask,
% for aesthetic reasons that you place these 'additional authors'
% in the \additional authors block, viz.
%% \additionalauthors{Additional authors: John Smith (The Th{\o}rv{\"a}ld Group,
%% email: {\texttt{jsmith@affiliation.org}}) and Julius P.~Kumquat
%% (The Kumquat Consortium, email: {\texttt{jpkumquat@consortium.net}}).}
%% \date{30 July 1999}
% Just remember to make sure that the TOTAL number of authors
% is the number that will appear on the first page PLUS the
% number that will appear in the \additionalauthors section.

\maketitle
\begin{abstract} 
In this paper we consider the problem of multi-view face detection. While there has been significant research on this problem, current state-of-the-art approaches for this task require annotation of facial landmarks, e.g. TSM \cite{tsm}, or  annotation of face poses \cite{multires_cascade, head_hunter}. They also require training dozens of models to fully capture faces in all orientations, e.g. $22$ models in HeadHunter method \cite{head_hunter}. In this paper we propose Deep Dense Face  Detector (DDFD), a method that does not require pose/landmark annotation and is able to detect faces in a wide range of orientations using a \textit{single} model based on deep convolutional neural networks. The proposed method has minimal complexity; unlike other recent deep learning object detection methods \cite{rcnn}, it does not require additional components such as segmentation, bounding-box regression, or SVM classifiers. Furthermore, we analyzed scores of the proposed face detector for faces in different orientations and found that 1) the proposed method is able to detect faces from different angles and can handle occlusion to some extent, 2) there seems to be a correlation between distribution of positive examples in the training set and scores of the proposed face detector. The latter suggests that the proposed method's performance can be further improved by using better sampling strategies and more sophisticated data augmentation techniques.    Evaluations on popular face detection benchmark datasets show that our single-model face detector algorithm has similar or better performance compared to the previous methods, which are more complex and require  annotations of either different poses or facial landmarks.
% * <fsachin@yahoo-inc.com> 2015-03-04T01:22:27.618Z:
%
% 
%
\end{abstract}

%A category including the fourth, optional field follows...
%\category{D.2.8}{Software Engineering}{Metrics}[complexity measures, performance measures]



\begin{figure}[t] 
  \centering
    \includegraphics[height=1.9in]{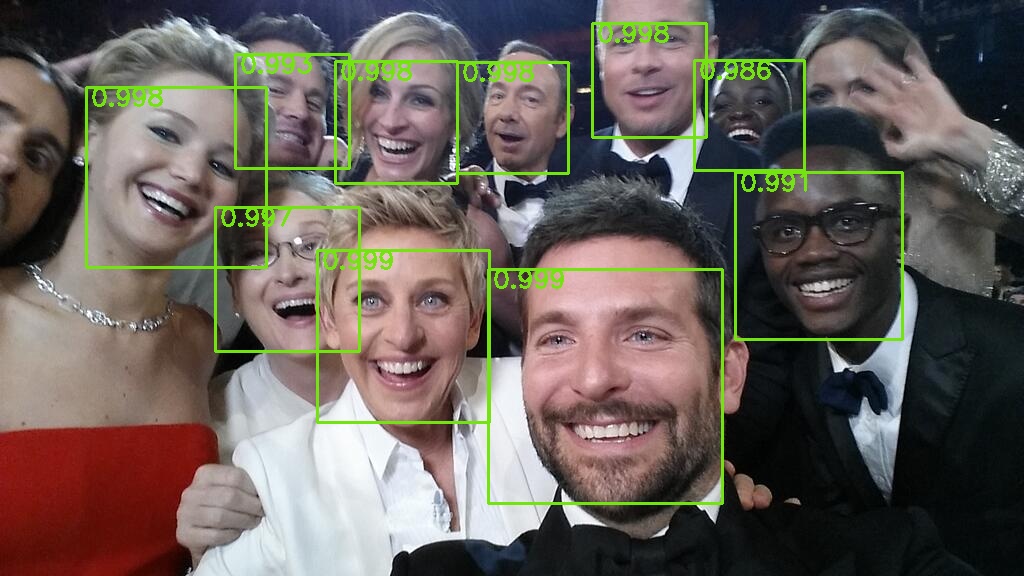}
  \caption{\protect{\footnotesize{An example of user generated photos on social networks that contains faces in various poses, illuminations and occlusions. The bounding-boxes and corresponding scores show output of our proposed face detector.}}}
  \label{fig:challenging_faces}
\end{figure}

\section{Introduction}

With the wide spread use of smartphones and fast mobile networks, millions of photos are uploaded everyday to the cloud storages such as Dropbox or social networks such as Facebook, Twitter, Instagram, Google+, and Flicker. Organizing and retrieving relevant information from these photos is very challenging and directly impact user experience on those platforms. For example, users commonly look for photos that were taken at a particular location, at a particular time, or with a particular friend. The former two queries are fairly straightforward, as almost all of today's cameras embed time and GPS location into photos. The last query, i.e. contextual query, is more challenging as there is no explicit signal about the identities of people in the photos. The key for this identification is the detection of human faces. This has made low complexity, rapid and accurate face detection an essential component for cloud based photo sharing/storage platforms.

For the past two decades, face detection has always been an active research area in the vision community. The seminal work of Viola and Jones \cite{VJ} made it possible to rapidly detect up-right faces in real-time with very low computational complexity. Their detector, called detector cascade, consists of a sequence of simple-to-complex face classifiers and has attracted extensive research efforts. Moreover, detector cascade has been deployed in many commercial products such as smartphones and digital cameras. 
While cascade detectors can accurately find visible up-right faces, they often fail to detect faces from different angles, e.g. side view or partially occluded faces. This failure can significantly impact the performance of photo organizing software/applications  since user generated content often contains faces from different angles or faces that are not fully visible; see for example Figure \ref{fig:challenging_faces}. This has motivated many works on the problem of multi-view face detection over the past two decades. Current solutions can be summarized into three categories: 
\begin{itemize}
\item{Cascade-based: These methods extend the Viola and Jones detector cascade. For example, \cite{parallel_cascade} proposed to train a detector cascade for each view of the face and combined their results at the test time. Recently, \cite{head_hunter} combined this method with integral channel features \cite{integral_channel} and soft-cascade \cite{soft_cascade}, and showed that by using $22$ cascades, it is possible to obtain state-of-the-art performance for multi-view face detection. This approach, however, requires face orientation annotations. Moreover its complexity in training and testing increases linearly with the number of models. To address the computational complexity issue, Viola and Jones \cite{VJ_multi_view} proposed to first estimate the face pose using a tree classifier and then run the cascade of corresponding face pose to verify the detection. While improving the detection speed, this method degrades the accuracy because mistakes of the initial tree classifier are irreversible. This method is further improved by \cite{rotation_invariant, vector_boost} where, instead of one detector cascade, several detectors are used after the initial classifier. Finally, \cite{torralba_cascade} and \cite{multires_cascade} combined detector cascade with multiclass boosting and proposed a method for multiclass/multi-view object detection.}
\item{DPM-based: These methods are based on the deformable part models technique \cite{dpm} where a face is defined as a collection of its parts. The parts are defined via unsupervised or supervised training, and a classifier, latent SVM, is trained to find those parts and their geometric relationship. These detectors are robust to partial occlusion because they
can detect faces even when some of the parts are not present. These methods are, however, computationally intensive because 1) they require solving a latent SVM for each candidate location and 2) multiple DPMs have to be trained and combined to achieve the state-of-the-art performance  \cite{head_hunter, tsm}. Moreover, in some cases DPM-based models require annotation of facial landmarks for training, e.g \cite{tsm}.}
\item{Neural-Network-based: There is a long history of using neural networks for the task of face detection \cite{Vaillant94, Vaillant93, Rowley98, Garcia04, Garcia03, Garcia02, embeddedCFF, openMPCFF, Osadchy04, Osadchy07}. In particular, \cite{Vaillant94} trained a two-stage system based on convolutional neural networks. The first network locates rough positions of faces and the second network verifies the detection and makes more accurate localization. In \cite{Rowley98}, the authors trained multiple face detection networks and combined their output to improve the performance. \cite{Garcia04} trained a single multi-layer network for face detection. The trained network is able to partially handle different poses and rotation angles. More recently, \cite{Osadchy07} proposed to train a neural network jointly for face detection and pose estimation. They showed that this joint learning scheme can significantly improve  performance of both detection and pose estimation. Our method follows the works in \cite{Garcia04, Osadchy07} but  constructs a deeper CNN for face detection.}

%Our approach to multi-view face detection is motivated by the success of those approaches using convolutional neural network.}%

%These  pioneering approaches of using convolutional neural network for face detection have %been inspiring to our paper.  In our experiment comparison, we focus on comparing our %approaches to the current state-of-the-art methods which are extensions of either the %detector cascade \cite{VJ} or the deformable part models \cite{dpm}.
%}
\end{itemize} 

The key challenge in multi-view face detection, as pointed out by Viola and Jones \cite{VJ_multi_view}, is that learning algorithms such as Boosting or SVM and image features such as HOG or  Haar wavelets are not strong enough to capture faces of different poses and thus the resulted classifiers are \textit{hopelessly inaccurate}. However, with recent advances in deep learning and GPU computation, it is possible to utilize the high capacity of deep convolutional neural networks for feature extraction/classification, and train a \textit{single} model for the task of multi-view face detection.

Deep convolutional neural network has recently demonstrated outstanding performance in a variety of vision tasks such as face recognition \cite{deepface, deepid2}, object classification \cite{alex-net, googlenet}, and object detection \cite{rcnn, overfeat, spp, google_detector}. In particular \cite{alex-net} trained an $8$-layered network, called AlexNet, and showed that deep convolutional neural networks can significantly outperform other methods for the task of large scale image classification. For the task of object detection, \cite{rcnn} proposed R-CNN method that uses an image segmentation technique,  selective search \cite{selective_search}, to find candidate image regions and classify those candidates using a version of AlexNet that is fine-tuned for objects in the PASCAL VOC dataset. More recently, \cite{google_2014} improved R-CNN by 1) augmenting the selective search proposals with candidate regions from multibox approach \cite{multi_box}, and 2) replacing 8-layered AlexNet with a much deeper CNN model of GoogLeNet \cite{googlenet}. Despite state-of-the-art performance, these methods are computationally sub-optimal because they require evaluating a CNN over more than $2,000$ overlapping candidate regions independently. To address this issue, \cite{spp} recently proposed to run the CNN model on the full image once and create a feature pyramid. The candidate regions, obtained by selective search, are then mapped into this feature pyramid space. \cite{spp} then uses spatial pyramid pooling \cite{spatial_pyramid_pooling} and SVM on the mapped regions to classify candidate proposals. Beyond region-based methods, deep convolutional neural networks have also been used with sliding window approach, e.g. OverFeat \cite{overfeat} and  deformable part models \cite{dense_net_dpm} for object detection and \cite{Tompson14} for human pose estimation. In general, for object detection these methods still have an inferior performance compared to region-based methods such as R-CNN \cite{rcnn} and \cite{google_2014}. However, in our face detection experiments we found that the region-based methods are often very slow and result in relatively weak performance. 

In this paper, we propose a method based on deep learning, called Deep Dense Face Detector (DDFD), that does not require pose/landmark annotation and is able to detect faces in  a wide range of orientations using a \textit{single} model. The proposed method has minimal complexity because unlike recent deep learning object detection methods such as \cite{rcnn}, it does not require additional components for segmentation, bounding-box regression, or SVM classifiers. Compared to previous convolutional neural-network-based face detectors such as \cite{Garcia04}, our network is deeper and is trained on a significantly larger training set. 
%We also analyzed the performance of our method on a variety of face images with different orientations and found that DDFD is able to detect faces in a wide range of orientations. 
In addition, by analyzing detection confidence scores, we show that there seems to be a correlation between the distribution of positive examples in the training set and the confidence scores of the proposed detector. This suggests that the performance of our method can be further improved by using better sampling strategies and more sophisticated data augmentation techniques. In our experiments, we compare the proposed method to a deep learning based method, R-CNN, and several cascade and DPM-based methods. We show that DDFD can achieve similar or better performance even without using pose annotation or information about facial landmarks.

% motivate the method:
% 	rccn is slow and can miss faces
% 	simplify the detector
%     single model CNN seems to be strong enough to capture the variations among all poses and occlusion

% 	going away from 
%     	selective search: 
%         SVM:
% 		bbox regression:

\begin{figure*}[t]
  \centering
	\begin{tabular}{cc}
        \includegraphics[height=2in, width=3.4in]{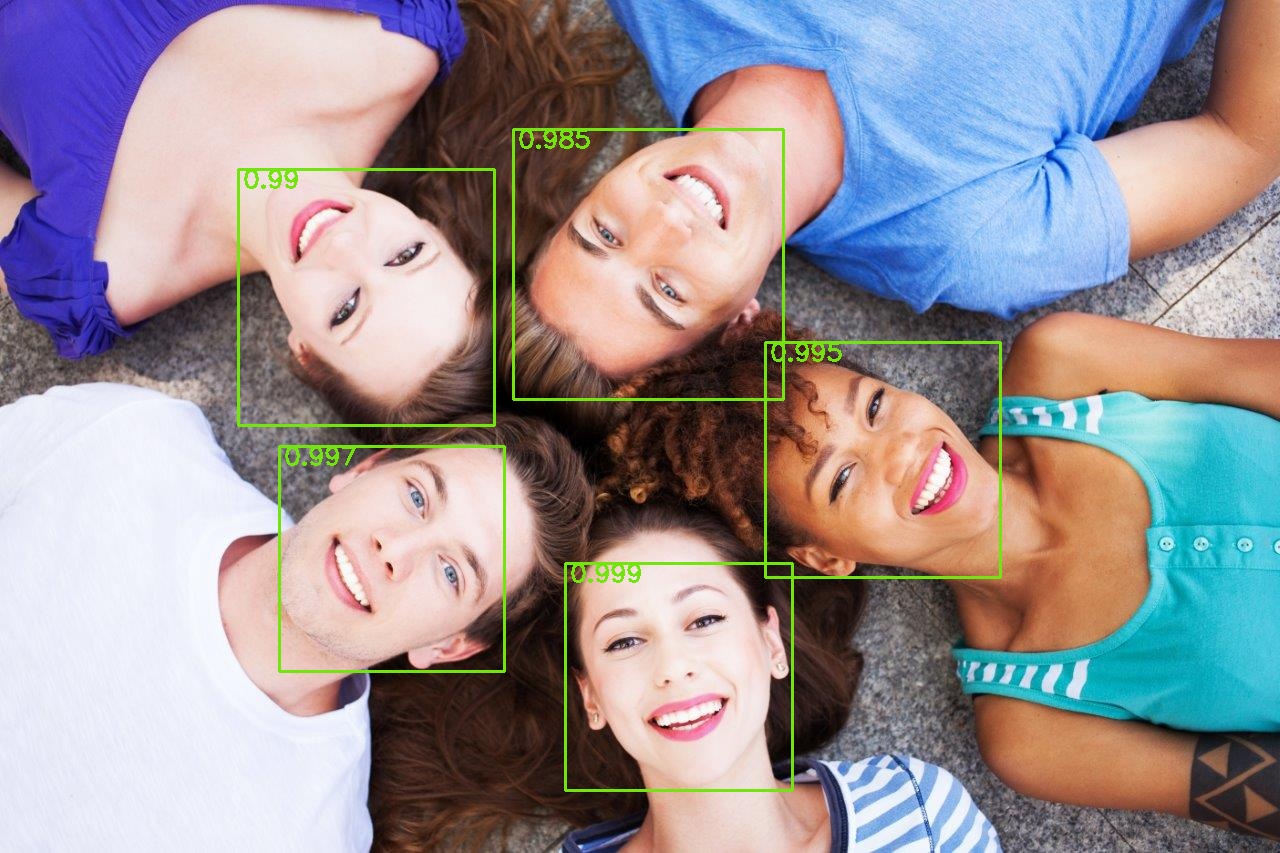} &
        \includegraphics[height=2in, width=3.4in]{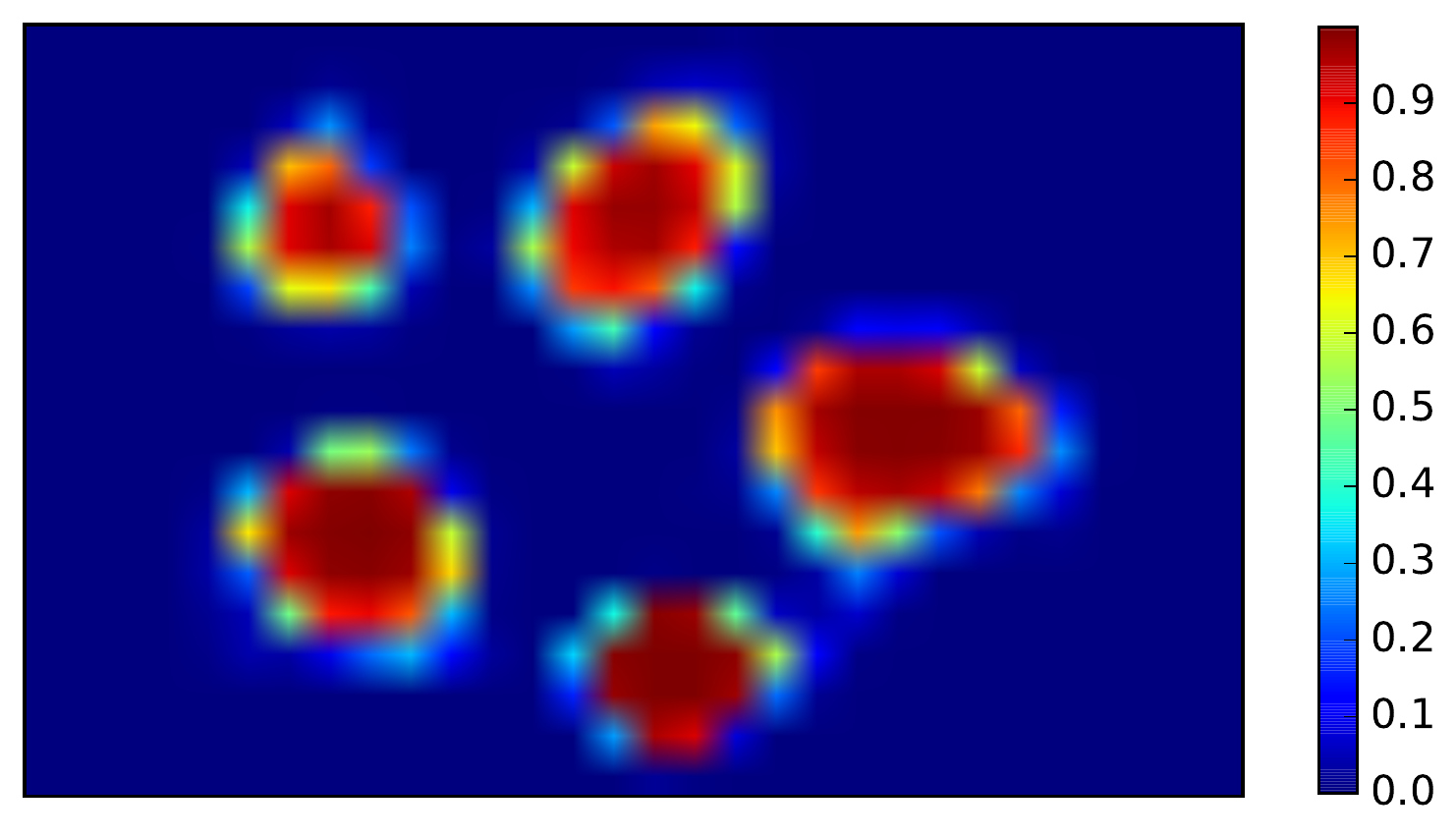}
    \end{tabular}
  \caption{\protect{\footnotesize{left) an example image with faces in different in-plane rotations. It also shows output of our proposed face detector after NMS along with corresponding confidence score for each detection. right) heat-map for the response of DDFD scores over the image.}}}
  \label{fig:heat_map}
  %\vspace{-.1in}
\end{figure*}

\section{Proposed Method}
\label{sec:proposed_method}
In this section, we provide details of the algorithm and training process of our proposed face detector, called Deep Dense Face Detector (DDFD). The key ideas are 1) leverage the high capacity of deep convolutional networks for classification and feature extraction to learn a single classifier for detecting faces from multiple views and 2) minimize the computational complexity by simplifying the architecture of the detector.

We start by fine-tuning AlexNet \cite{alex-net} for face detection. For this  we extracted training examples from the AFLW dataset \cite{AFLW}, which consists of $21$K images with $24$K face annotations. To increase the number of positive examples, we randomly sampled sub-windows of the images and used them as positive examples if they had more than a $50\%$ IOU (intersection over union) with the ground truth. For further data augmentation, we also randomly flipped these training examples. This resulted in a total number of $200$K positive and and $20$ millions negative training examples. These examples were then resized to $227\times227$ and used to fine-tune a pre-trained AlexNet model \cite{alex-net}. 
For fine-tuning, we used $50$K iterations and batch size of $128$ images, where each batch contained $32$ positive and $96$ negative examples. 

Using this fine-tuned deep network, it is possible to take either region-based or sliding window approaches to obtain the final face detector. In this work we selected a sliding window approach because it has less complexity and is independent of extra modules such as selective search. Also, as discussed in the experiment section, this approach leads to better results as compared to R-CNN.

%Using a deep CNN in sliding window is not possible for a network that has fully-connected layers. 

Our face classifier, similar to AlexNet \cite{alex-net}, consists of $8$ layers where the first $5$ layers are convolutional and the last $3$ layers are fully-connected. We first converted the fully-connected layers into convolutional layers by reshaping layer parameters \cite{densenet}. This made it possible to efficiently run the CNN on images of any size and obtain a heat-map of the face classifier. An example of a heat-map is shown in Figure \ref{fig:heat_map}-right. Each point in the heat-map shows the CNN response, the probability of having a face, for its corresponding $227 \times 227$ region in the original image. The detected regions were then processed by non-maximal suppression to accurately localize the faces. Finally, to detect faces of different sizes, we scaled the images up/down and obtained new heat-maps. We tried different scaling schemes and found that rescaling image $3$ times per octave gives reasonably good performance. This is interesting as many of the other methods such as \cite{head_hunter, pyramid_piotr} requires a significantly larger number of resizing per octave, e.g. $8$. Note that, unlike R-CNN \cite{rcnn}, which uses SVM classifier to obtain the final score, we removed the SVM module and found that the network output are informative enough for the task of face detection.

 Face localization can be further improved by using a bounding-box regression module similar to \cite{overfeat, rcnn}. In our experiment, however, adding this module degraded the performance. Therefore, compared to the other methods such as R-CNN \cite{rcnn}, which uses selective search, SVM and bounding-box regression, or DenseNet \cite{dense_net_dpm}, which is based on the deformable part models, our proposed method (DDFD) is fairly simple. Despite its simplicity, as shown in the experiments section, DDFD can achieve state-of-the-art performance for face detection.

\subsection{Detector Analysis}
        
In this section, we look into the scores of the proposed face detector and observe that
there seems to be a correlation between those scores and the distribution of positive examples in the training set. We can later use this hypothesis to obtain better training set or to design better data augmentation procedures and improve performance of DDFD.

\begin{figure}[t]
  \centering
	\begin{tabular}{c}
    	\includegraphics[width=3.2in]{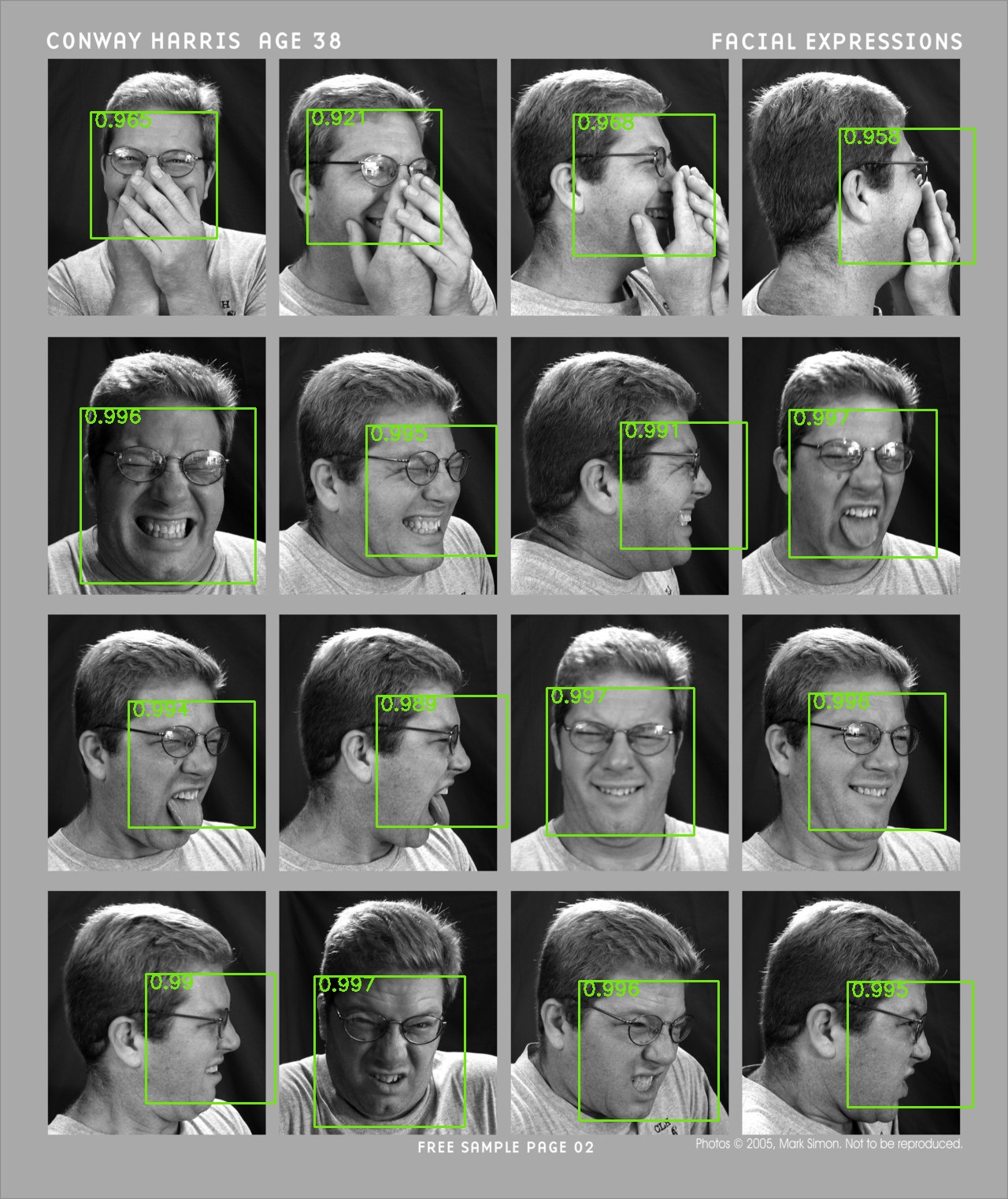} \\
    \end{tabular}
  \caption{\protect{ A set of faces with different out-of-plane rotations and occlusions. The figure also shows output of our proposed face detector after NMS along with the corresponding confidence score for each detection.}}
  \label{fig:face_rotations}
  %\vspace{-.1in}
\end{figure}

We begin by running our detector on a variety of faces with different in-plane and out-of-plane rotations, occlusions and lighting conditions (see for example  Figure \ref{fig:challenging_faces}, Figure \ref{fig:heat_map}-left and Figure \ref{fig:face_rotations}). First, note that in all cases our detector is able to detect the faces except for the two highly occluded ones in Figure \ref{fig:challenging_faces}. Second, for almost all of the detected faces, the detector's confidence score is pretty high, close to $1$. Also as shown in the heat-map of Figure \ref{fig:heat_map}-right, the scores are close to zero for all other regions. This shows that DDFD has very strong discriminative power, and its output can be used directly without any post-processing steps such as SVM, which is used in R-CNN \cite{rcnn}. Third, if we compare the detector scores for faces in Figure \ref{fig:heat_map}-left, it is clear that the up-right frontal face in the bottom has a very high score of $0.999$ while faces with more in-plane rotation have less score. Note that these scores are output of a sigmoid function, i.e. probability (soft-max) layer in the CNN, and thus small changes in them reflects much larger changes in the output of the previous layer. It is interesting to see that the scores decrease as the in-plane rotation increases. We can see the same trend for out-of-plane rotated faces and occluded faces in Figures \ref{fig:challenging_faces} and \ref{fig:face_rotations}. We hypothesize that this trend in the scores is not because detecting rotated face are more difficult but it is because of lack of good training examples to represent such faces in the training process.

\begin{figure}[t]
  \centering
	\begin{tabular}{c}
    	\includegraphics[height=1.5in]{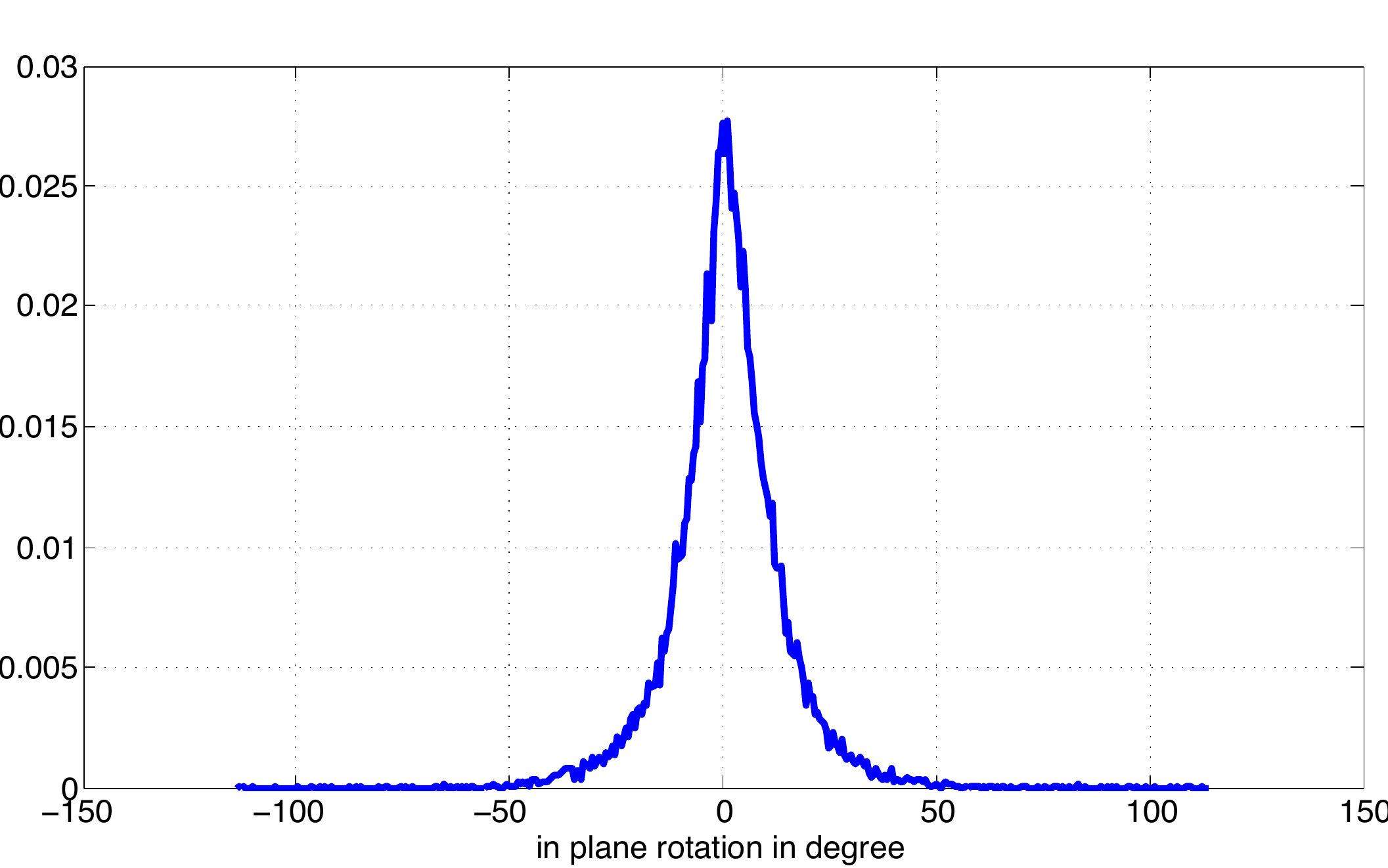} \\
    	\includegraphics[height=1.5in]{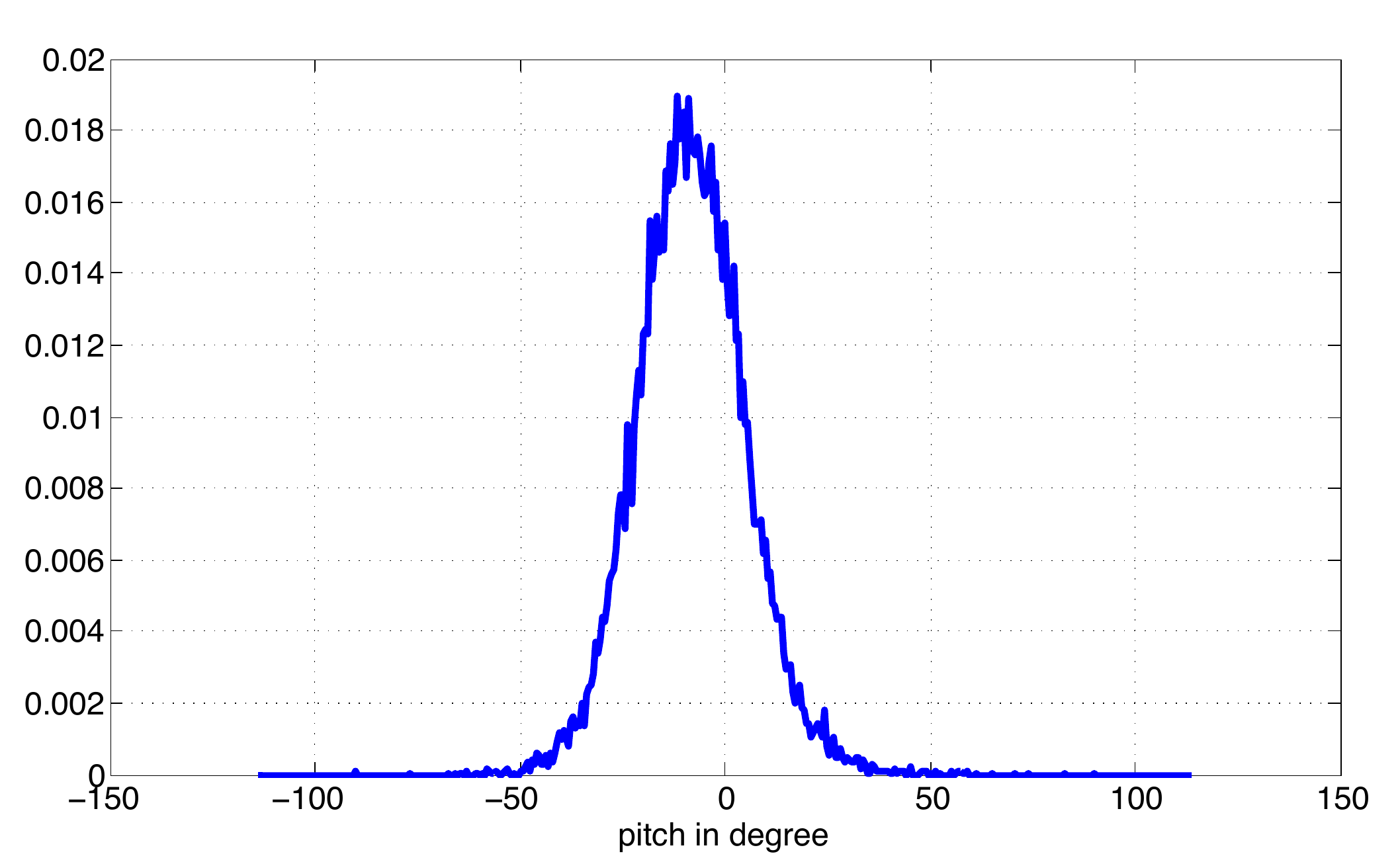} \\
    	\includegraphics[height=1.5in]{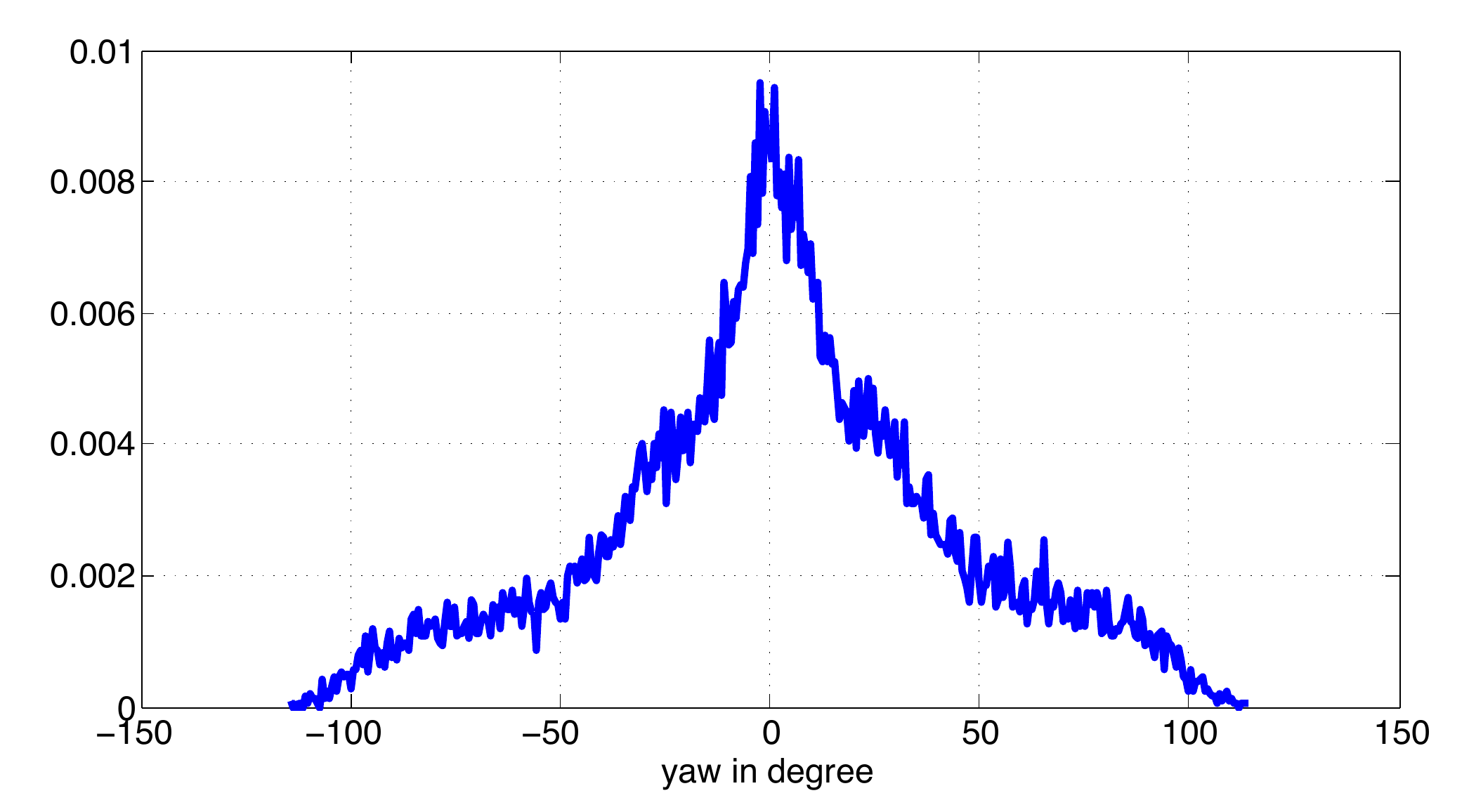} \\
    \end{tabular}
  \caption{\protect{ Histogram of faces in AFLW dataset based on their top) in-plane, middle) pitch (up and down) and bottom) yaw(left to right) rotations.}}
  \label{fig:aflw_example_hist}
  %\vspace{-.1in}
\end{figure}

To examine this hypothesis, we looked into the face annotations for AFLW dataset \cite{AFLW}. Figure \ref{fig:aflw_example_hist} shows the distribution of the annotated faces with regards to their in-plane, pitch (up and down) and yaw (left to right) rotations. As shown in this figure, the number of faces with more than $30$ degrees out-of-plane rotation is significantly lower than the faces with less than $30$ degree rotation. Similarly, the number of faces with yaw or pitch less than $50$ degree is significantly larger than the other ones. Given this skewed training set, it not surprising that the fine-tuned CNN is more confident about up-right faces. This is because the CNN is trained to minimize the risk of the soft-max loss function
\begin{equation}
{\cal R} = \sum_{x_i \in {\cal B}} \log\left[prob(y_i| x_i )\right],
\end{equation}
where ${\cal B}$ is the example batch that is used in an iteration of stochastic gradient descent and  $y_i$ is the label of example $x_i$. The sampling method for selecting examples in ${\cal B}$ can significantly hurt performance of the final detector. In an extreme case if ${\cal B}$ never contains any example of a certain class, the CNN classifier will never learn the attributes of that class.

In our implementation $|{\cal B}| = 128$ and it is collected by randomly sampling the training set. However, since the number of negative examples are $100$ times more than the number of positive examples, a uniform sampling will result in only about $2$ positive examples per batch. This significantly degrades the chance of the CNN to distinguish faces from non-faces. To address this issue, we enforced one quarter of each batch to be positive examples, where the positive examples are uniformly sampled from the pool of positive training samples. But, as illustrated in Figure \ref{fig:aflw_example_hist}, this pool is highly skewed in different aspects, e.g. in-plane and out-of-plane rotations. The CNN is therefore getting exposed with more up-right faces; it is thus not surprising that the fine-tuned CNN is more confident about the up-right faces than the rotated ones. This analysis suggests that the key for improving performance of DDFD is to ensure that all categories of the training examples have similar chances to contribute in optimizing the CNN. This can be accomplished by enforcing population-based sampling strategies such as increasing selection probability for categories with low population.
%similar to our strategy for sampling more positive examples.

Similarly, as shown in Figure \ref{fig:challenging_faces}, the current face detector still fails to detect faces with heavy occlusions. Similar to the issue with rotated faces, we believe that this problem can also be addressed through modification of the training set. In fact, most of the face images in the AFLW dataset \cite{AFLW} are not occluded, which makes it difficult for a CNN to learn that faces can be occluded. This issue can be addressed by using more sophisticated data augmentation techniques such as occluding parts of positive examples. Note that simply covering parts of positive examples with black/white or noise blocks is not useful as the CNN may learn those artificial patterns. 

To summarize, the proposed face detector based on deep CNN is able to detect faces from different angles and handle occlusion to some extent. However, since the training set is skewed, the network is more confident about up-right faces and better results can be achieved by using better sampling strategies and more sophisticated data augmentation techniques.

%#\section{Adjusting for dataset biases}
%\subsection{Bias in annotations}
%\subsection{Bias in training examples}
\section{Experiments}

We implemented the proposed face detector using the Caffe library \cite{caffe} and used its pre-trained Alexnet \cite{alex-net} model for fine-tuning. For further details on the training process of our proposed face detector please see section \ref{sec:proposed_method}. After converting fully-connected layers to convolutional layers \cite{densenet}, it is possible to get the network response (heat-map) for the whole input image in one call to Caffe code.  The heat-map shows the scores of the CNN for every $227 \times 227$ window with a stride of $32$ pixels in the original image. We directly used this response for classifying a window as face or background. To detect faces of smaller or larger than $227 \times 227$, we scaled the image up or down respectively.
   
We tested our face detection approach on PASCAL Face \cite{structural_model}, AFW \cite{tsm} and FDDB \cite{fddb} datasets. For selecting and tuning parameters of the proposed face detector we used the PASCAL Face dataset.
%results are finally reported on AFW and FDDB datasets as well. 
PASCAL Face dataset consists of $851$ images and $1341$ annotated faces, where annotated faces can be as small as $35$ pixels. AFW dataset is built using Flickr images. It has $205$ images with $473$ annotated faces, and its images tend to contain cluttered background with large variations in both face viewpoint and appearance (aging, sunglasses, make-ups, skin color, expression etc.). Similarly, FDDB dataset \cite{fddb} consists of $5171$ annotated faces with $2846$ images and contains occluded, out-of-focus, and low resolution faces. For evaluation, we used the toolbox provide by \cite{head_hunter} with corrected annotations for PASCAL Face and AFW datasets and the original annotations of FDDB dataset.

\begin{figure}[t]
  \centering
    \includegraphics[width=\columnwidth]{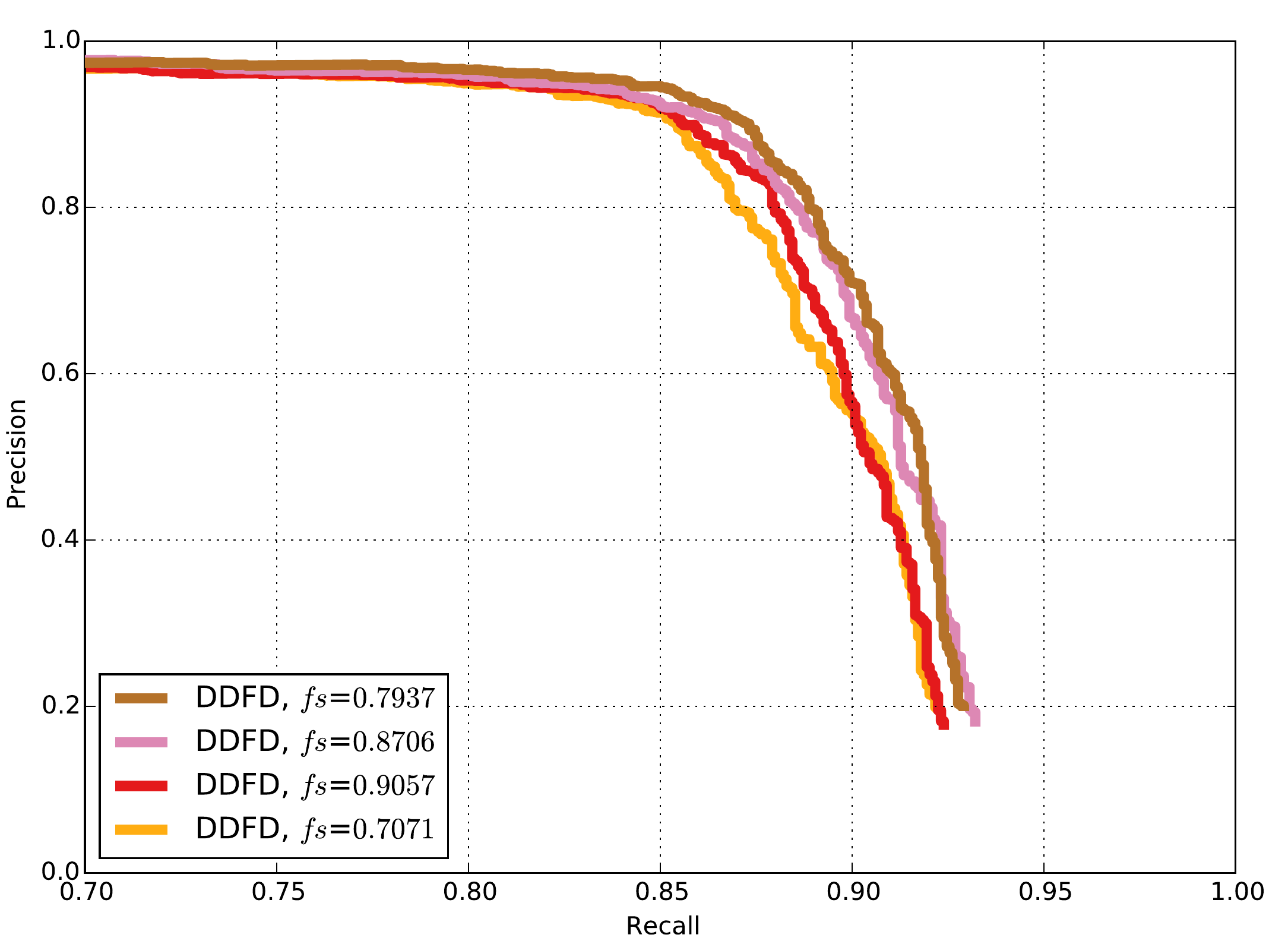}
  \caption{\protect{\footnotesize{Effect of scaling factor on precision and recall of the detector.}}}
  \label{fig:f_s_effect}
  %\vspace{-.1in}
\end{figure}

We started by finding the optimal number of scales for the proposed detector using PASCAL dataset. We upscaled images by factor of $5$ to detect faces as small as $227/5=45$ pixels. We then down scaled the image with by a factor, $f_s$, and repeated the process until the minimum image dimension is less than $227$ pixels. For the choice of $f_s$, we chose $f_s \in \{ \sqrt{0.5}=0.7071, \sqrt[3]{0.5}=0.7937,  \sqrt[5]{0.5}=0.8706, \sqrt[7]{0.5}=0.9056 \}$; Figure \ref{fig:f_s_effect} shows the effect of this parameter on the precision and recall of our face detector (DDFD). 
%The legend shows the area under the curve for each method. 
Decreasing $f_s$ allows the detector to scan the image finer and increases the computational time. According to Figure \ref{fig:f_s_effect}, it seems that these choices of $f_s$
%$f_s \in \{ \sqrt{0.5}=0.7071, \sqrt[3]{0.5}=0.7937,  \sqrt[5]{0.5}=0.8706, \sqrt[7]{0.5}=0.9056 \}$
has little impact on the performance of the detector. Surprisingly, $f_s =  \sqrt[3]{0.5}$ seems to have slightly better performance although it does not scan the image as thorough as $f_s =  \sqrt[5]{0.5}$ or $f_s =\sqrt[7]{0.5}$. Based on this experiment we use $f_s =\sqrt[3]{0.5}$ for the rest of this paper.

\begin{figure}[t]
  \centering
    \includegraphics[width=\columnwidth]{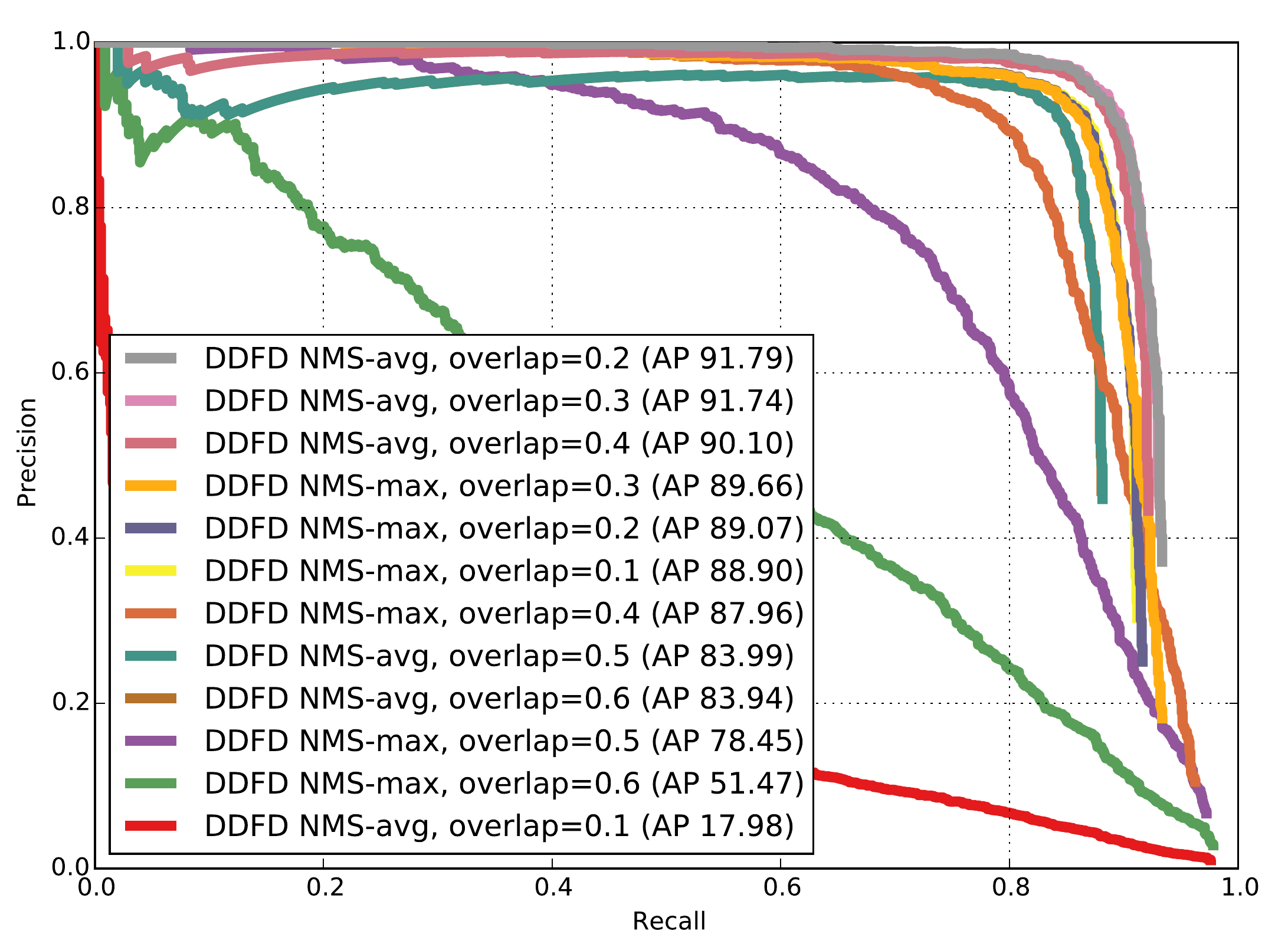}
  \caption{\protect{\footnotesize{Effect of different NMS strategies and their overlap thresholds.}}}
  \label{fig:f_nms_effect}
  %\vspace{-.1in}
\end{figure}

Another component of our system is the non-maximum suppression module (NMS). For this we evaluated two different strategies: 
\begin{itemize}
\item{\textbf{NMS-max}: we find the window of the maximum score and remove all of the bounding-boxes with an IOU (intersection over union) larger than an overlap threshold.}
\item{\textbf{NMS-avg}: we first filter out windows with confidence lower than $0.2$. We then use groupRectangles function of OpenCV to cluster the detected windows according to an overlap threshold. Within each cluster, we then removed all windows with score less than $90\%$ of the maximum score of that cluster. Next we averaged the locations of the remaining bounding-boxes to get the detection window. Finally, we used the maximum score of the cluster as the score of the proposed detection.}
\end{itemize}
We tested both strategies and Figure \ref{fig:f_nms_effect} shows the performance of each strategy for different overlap thresholds. As shown in this figure, performance of both methods vary significantly with the overlap threshold. An overlap threshold of $0.3$ gives the best performance for NMS-max while, for NMS-avg $0.2$ performs the best. According to this figure, NMS-avg has better performance compared to NMS-max in terms of average precision. 
%In the rest of this paper we use NMS-avg to show performance of our method.

\begin{figure}[t]
  \centering
    \includegraphics[width=\columnwidth]{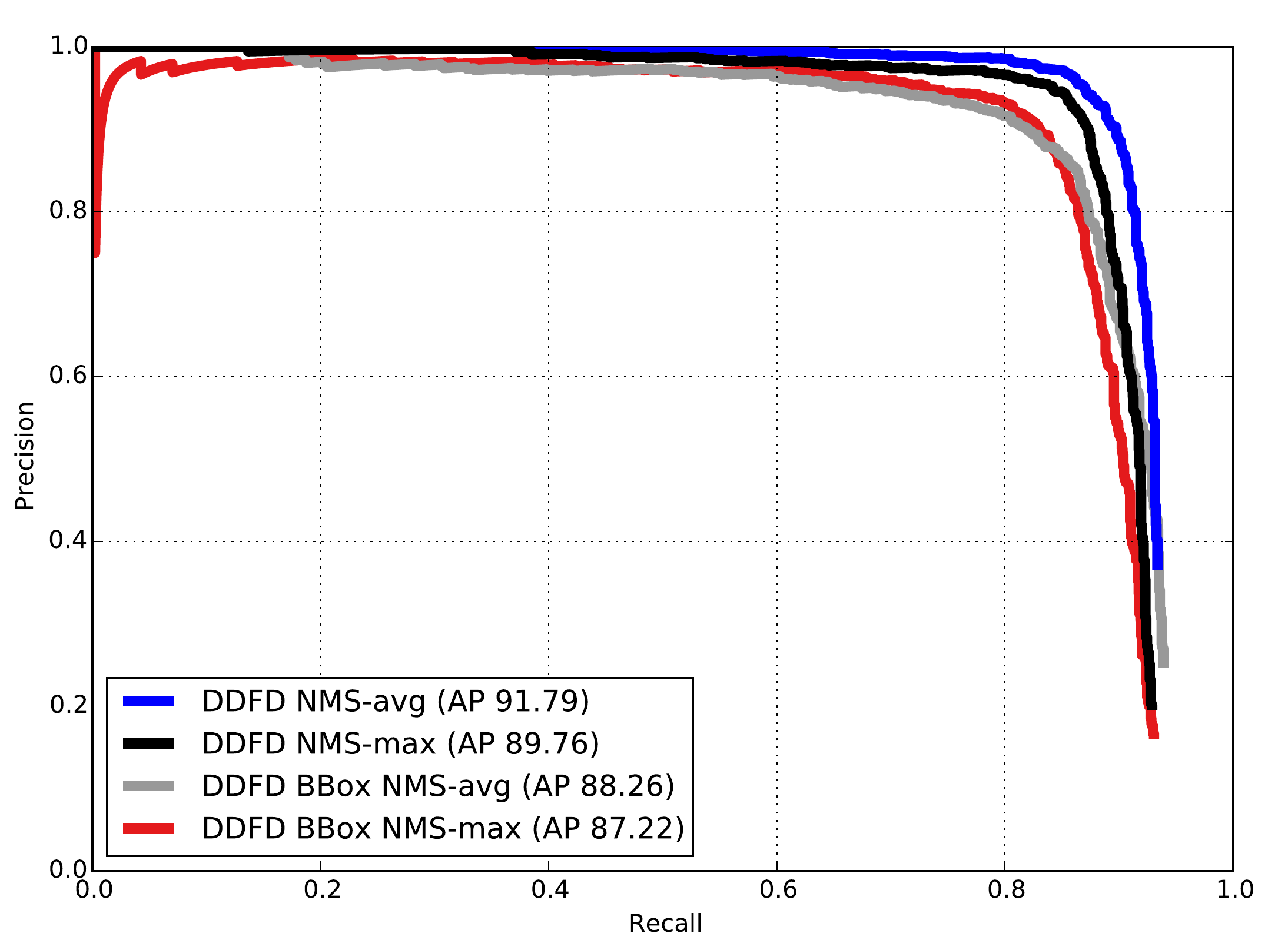}
  \caption{\protect{\footnotesize{Performance of the proposed face detector with and without bounding-box regression.}}}
  \label{fig:f_bbx_reg_effect}
  %\vspace{-.1in}
\end{figure}

\begin{figure}[t]
  \centering
  	\begin{tabular}{cc}
    	\includegraphics[height=1.5in]{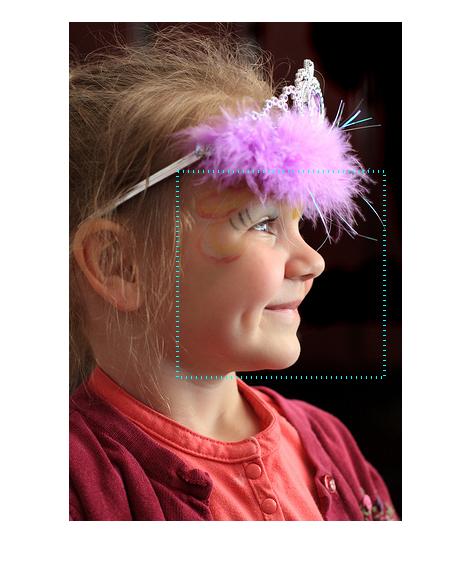} &
    	\includegraphics[height=1.5in]{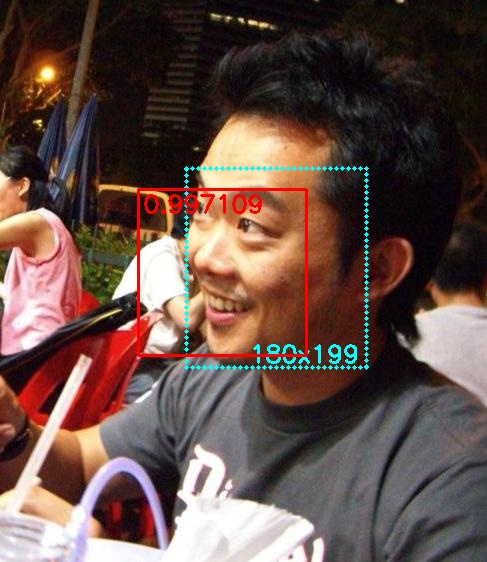} 
    \end{tabular}
  \caption{\protect{\footnotesize{Annotation of a side face in left) training set and right) test set. The red bounding-box is the predicted bounding-box by our proposed detector. This detection is counted as a false positive as its IOU with ground truth is less than $50\%$.}}}
  \label{fig:train_test_mismatch}
  %\vspace{-.1in}
\end{figure}

Finally, we examined the effect of a bounding-box regression module for improving detector localization. The idea is to train regressors to predict the difference between the locations of the predicted bounding-box and the ground truth. At the test time these regressors can be used to estimate the location difference and adjust the predicted bounding-boxes accordingly. This idea has been shown to improve localization performance in several methods including \cite{dpm, overfeat, multi_box}. To train our bounding-box regressors, we followed the algorithm of \cite{rcnn} and Figure \ref{fig:f_bbx_reg_effect} shows the performance of our detector with and without this module. As shown in this figure, surprisingly, adding a bounding-box regressors degrades the performance for both NMS strategies. Our analysis revealed that this is due to the mismatch between the annotations of the training set and the test set. This mismatch is mostly for side-view faces and is illustrated in Figure \ref{fig:train_test_mismatch}. In addition to degrading performance of bounding-box regression module, this mismatch also leads to false miss-detections in the evaluation process.

\subsection{Comparison with R-CNN} 

R-CNN \cite{rcnn} is one of the current state-of-the-art methods for object detection. In this section we compare our proposed detector with R-CNN and its variants.

\begin{figure}[t]
  \centering
    \includegraphics[height=2.4in, width=3.4in]{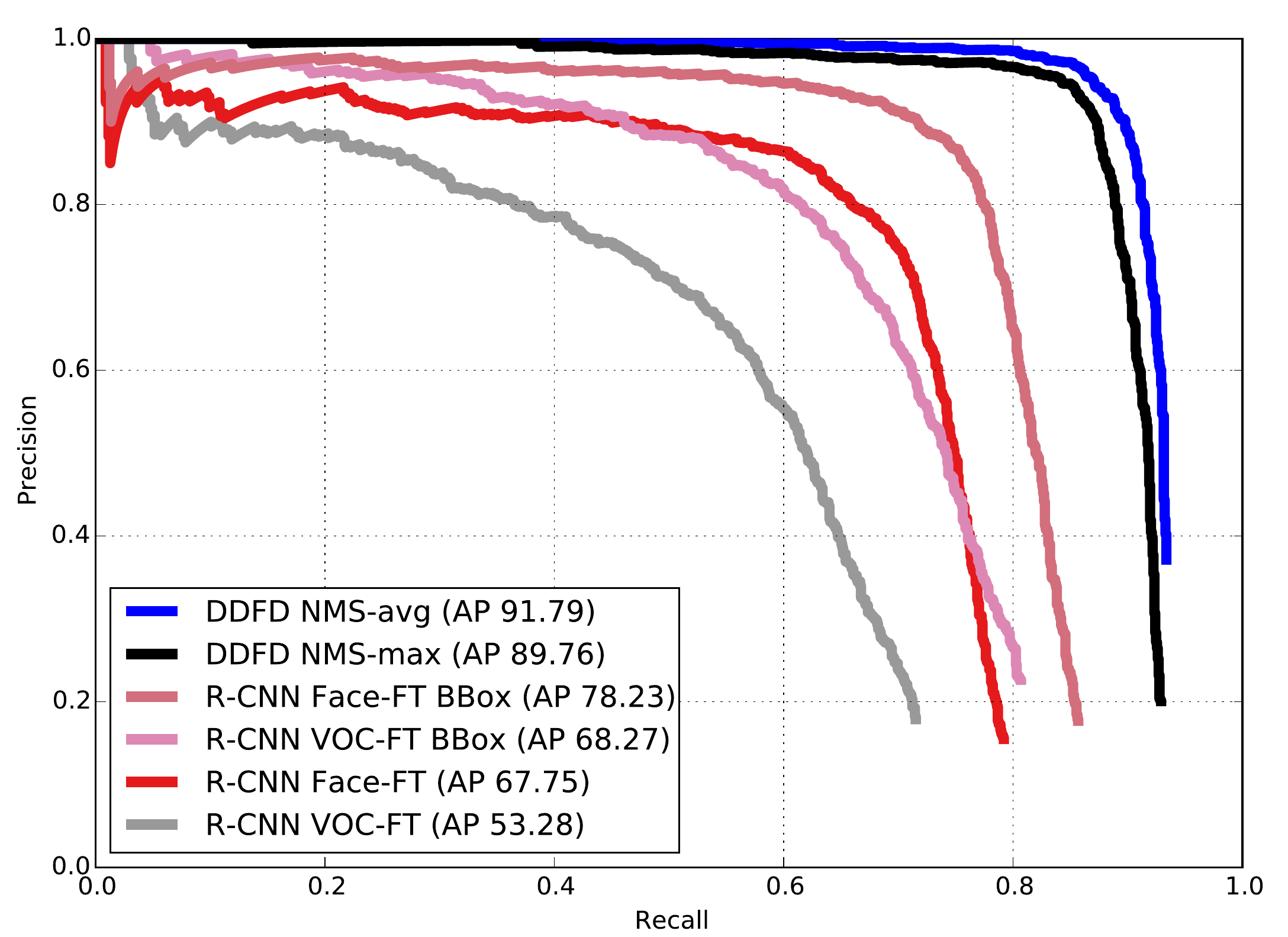}
  \caption{\protect{\footnotesize{Comparison of our face detector, DDFD, with different R-CNN face detectors.}}} 
  \label{fig:rcnn_results} 
  \vspace{-.1in}
\end{figure}

We started by fine-tuning AlexNet for face detection using the process described in section \ref{sec:proposed_method}. We then trained a SVM classifier for face classification using output of the seventh layer  ($fc7$ features). We also trained a bounding-box regression unit to further improve the results and used NMS-max for final localization. We repeated this experiment on a version of AlexNet that is fine-tuned for PASCAL VOC 2012 dataset and is provided with R-CNN code. Figure \ref{fig:rcnn_results} compares the performance of our detector with different NMS strategies along with the performance of R-CNN methods with and without bounding-box regression. As shown in this figure, it is not surprising that performance of the detectors with
AlexNet fine-tuned for faces (Face-FT) are better than the ones that are fine-tuned with PASCAL-VOC objects (VOC-FT). In addition, it seems that bounding-box regression can significantly improve R-CNN performance. However, even the best R-CNN classifier has significantly inferior performance compared to our proposed face detector independent of the NMS strategy. We believe the inferior performance of R-CNN are due to 1) the loss of recall since selective search may miss some of face regions and 2) loss in localization since bounding-box regression is not perfect and may not be able to fully align the segmentation bounding-boxes, provided by selective search \cite{selective_search}, with the ground truth.

\subsection{Comparisons with state-of-the-art} 

\begin{figure*}[t]
  \centering
  	\begin{tabular}{cc}
    	\includegraphics[height=2.4in, width=3.4in]{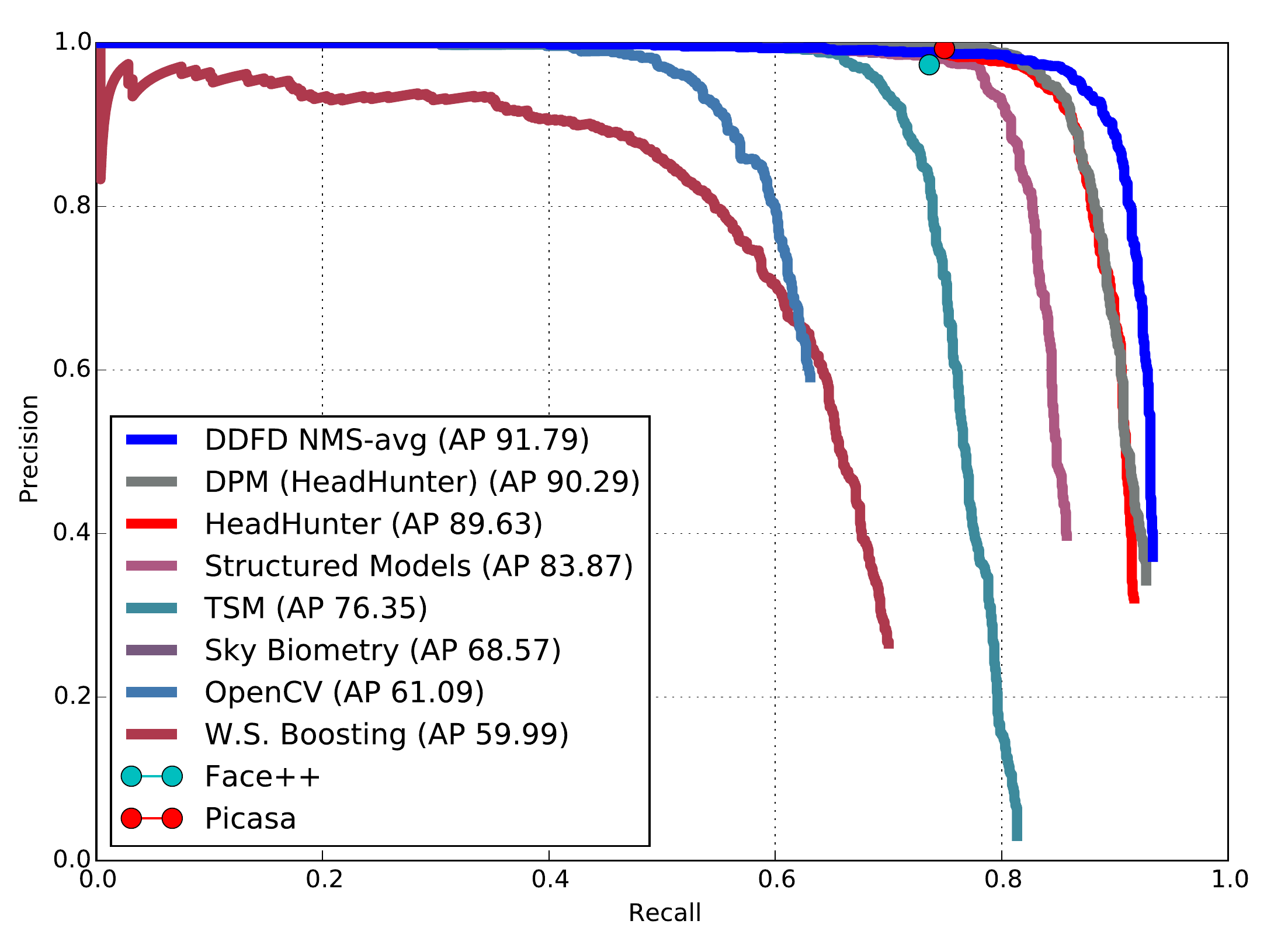} &
    	\includegraphics[height=2.4in, width=3.4in]{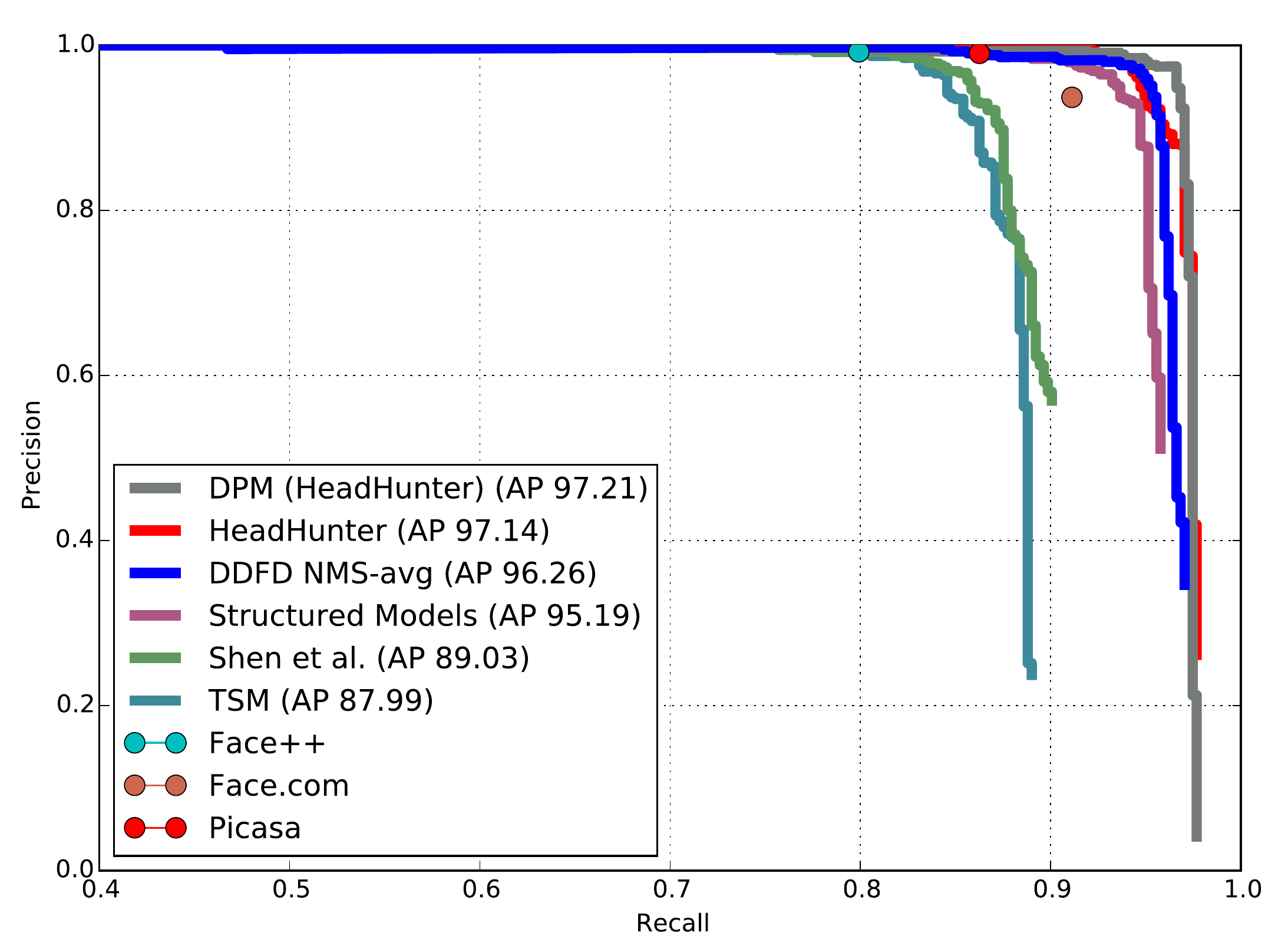}
    \end{tabular}
  \caption{\protect{\footnotesize{Comparison of different face detectors on left) PASCAL faces and right) AFW dataset.}}}
  \label{fig:performance comparison}
  \vspace{-.1in}
\end{figure*}

\begin{figure}[t]
  \centering
  \includegraphics[height=2.4in, width=3.4in]{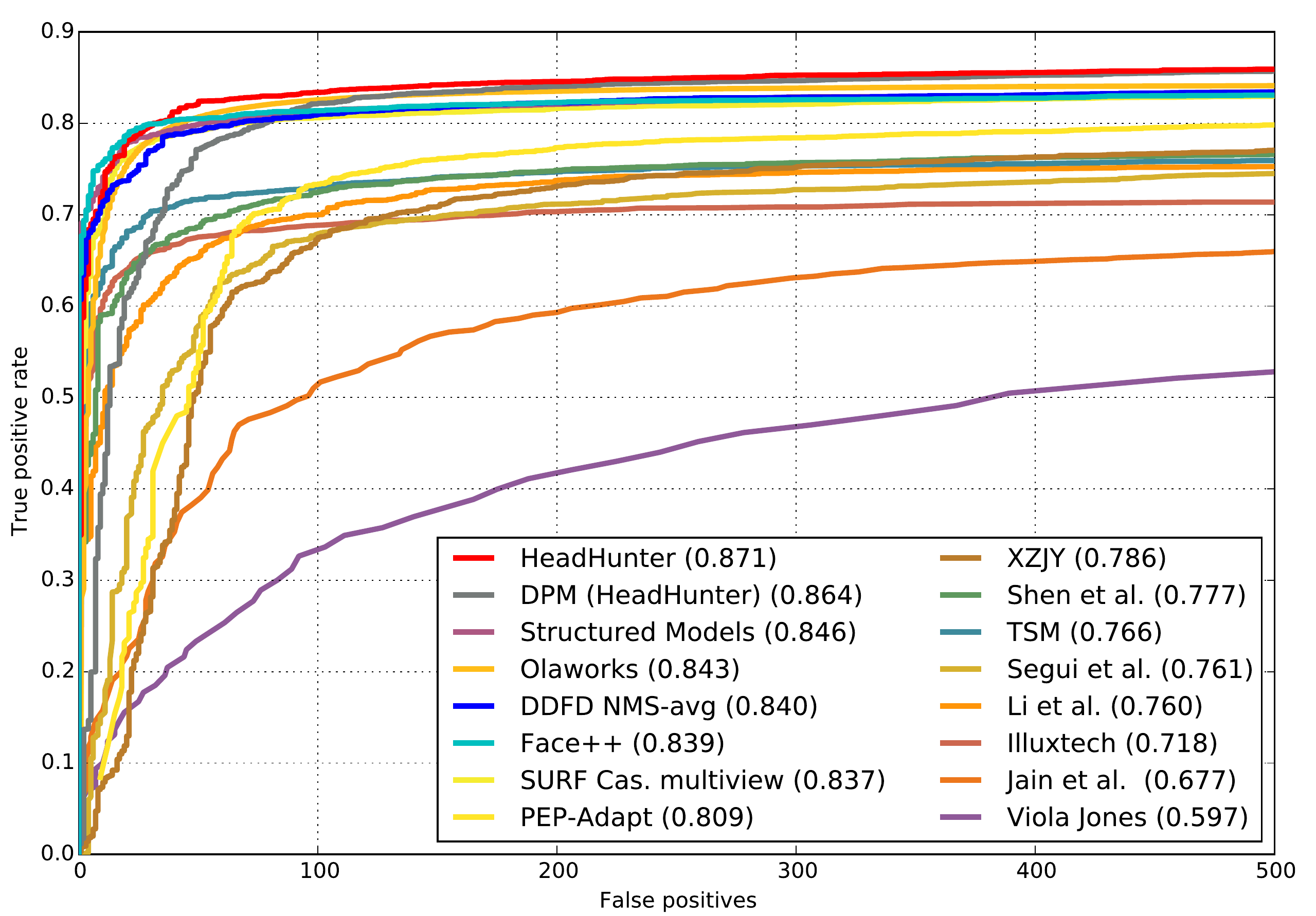}
  \caption{\protect{\footnotesize{Comparison of different face detectors on FDDB dataset.}}}
  \label{fig:performance_comparison_fddb}
    \vspace{-.1in}
\end{figure}

In this section we compare the performance of our proposed detector with other state-of-the-art face detectors using publicly available datasets of  PASCAL faces \cite{structural_model}, AFW \cite{tsm} and FDDB \cite{fddb}. In particular, we compare our method with 1) DPM-based methods such as structural model \cite{structural_model} and TSM \cite{tsm} and 2) cascade-based method such as head hunter \cite{head_hunter}. Figures \ref{fig:performance comparison} and \ref{fig:performance_comparison_fddb} illustrate this comparison. Note that these comparison are not completely fair as most of the other methods such as DPM or HeadHunter use extra information of view point annotation during the training. As shown in these figures our single model face detector was able to achieve similar or better results compared to the other state-of-the-art methods, without using pose annotation or information about facial landmarks.

%( selective search + fine tune for VOC + SVM on fc7 + bbx regression on pool5 + nms)
%

%pascal dataset 1341 faces (851 images)
%AFW dataset 473 (205 images)
%AFLW 24386 faces 
%(PASCAL + FDDB discrete and continues + less 80 + larger than 80 AFW) curves on three datasets         
%Add run time for different datasets.
%discussion about results 
%  good results on rotation and occlusion
%  illustrate some of the mistakes
%  some future works

\section{Conclusions and future work}
In this paper, we proposed a face detection method based on deep learning, called Deep Dense Face Detector (DDFD).  The proposed method does not require pose/landmark annotation and is able to detect faces in a wide range of orientations using a \textit{single} model. In addition, DDFD is independent of common modules in recent deep learning object detection methods such as bounding-box regression, SVM, or image segmentation. We compared the proposed method with R-CNN and  other face detection methods that are developed specifically for multi-view face detection e.g. cascade-based and DPM-based. We showed that our detector is able to achieve similar or better results even without using pose annotation or information about facial landmarks. Finally, we analyzed performance of our proposed face detector on a variety of face images and found that there seems to be a correlation between distribution of positive examples in the training set and scores of the proposed detector. In future we are planning to use better sampling strategies and more sophisticated data augmentation techniques to further improve performance of the proposed method for detecting occluded and rotated faces. 

%\begin{figure*}[t]
%  \centering
%  	\begin{tabular}{cc}
%    	\includegraphics[width=3.3in, height=2.1in]{figs/4856974482.jpg} &    	%\includegraphics[width=3.3in, height=2.1in]{figs/4400753623.jpg} \\
   %     \includegraphics[height=1.6in]{figs/4022732812.jpg} &
   % 	\includegraphics[width=3.3in, height=1.6in]{figs/4538917191.jpg} 
%        \includegraphics[width=3.3in, height=2.3in]{figs/5169701116.jpg} &
%    	\includegraphics[width=3.3in, height=2.3in]{figs/18489332_det.jpg} \\
%        \includegraphics[width=3.3in, height=2.3in]{figs/14424972.jpg} &    	\includegraphics[width=3.3in, height=2.3in]{figs/213654866.jpg} 
%    \end{tabular}
%  \caption{\protect{\footnotesize{Examples of output of our face detector from  AFW dataset. The results are shown in different colors: green boxes are true positives, magenta boxes are ground truth, red boxes are false positives and cyan boxes are false negatives.}}}
%  \label{fig:example_detections}
%\end{figure*}

\bibliographystyle{abbrv}
\bibliography{sigproc}

\begin{thebibliography}{10}

\bibitem{soft_cascade}
L.~Bourdev and J.~Brandt.
\newblock Robust object detection via soft cascade.
\newblock In {\em Proceedings of CVPR}, 2005.

\bibitem{pyramid_piotr}
P.~Doll\'ar, R.~Appel, S.~Belongie, and P.~Perona.
\newblock Fast feature pyramids for object detection.
\newblock {\em IEEE Transactions on Pattern Analysis and Machine Intelligence},
  2014.

\bibitem{integral_channel}
P.~Dollar, Z.~Tu, P.~Perona, and S.~Belongie.
\newblock Integral channel features.
\newblock In {\em Proceedings of the British Machine Vision Conference}, 2009.

\bibitem{multi_box}
D.~Erhan, C.~Szegedy, A.~Toshev, and D.~Anguelov.
\newblock Scalable object detection using deep neural networks.
\newblock 2014.

\bibitem{dpm}
P.~Felzenszwalb, D.~McAllester, and D.~Ramanan.
\newblock A discriminatively trained, multiscale, deformable part model.
\newblock In {\em Proceedings of CVPR}, 2008.

\bibitem{Garcia02}
C.~{Garcia} and M.~{Delakis}.
\newblock {A Neural Architecture for Fast and Robust Face Detection}.
\newblock In {\em Proceedings of IEEE-IAPR International Conference on Pattern
  Recognition}, Aug. 2002.

\bibitem{Garcia03}
C.~{Garcia} and M.~{Delakis}.
\newblock {Training Convolutional Filters for Robust Face Detection}.
\newblock In {\em Proceedings of IEEE International Workshop of Neural Networks
  for Signal Processing}, Sept. 2003.

\bibitem{Garcia04}
C.~Garcia and M.~Delakis.
\newblock Convolutional face finder: a neural architecture for fast and robust
  face detection.
\newblock {\em IEEE Transactions on Pattern Analysis and Machine Intelligence},
  2004.

\bibitem{rcnn}
R.~Girshick, J.~Donahue, T.~Darrell, and J.~Malik.
\newblock Rich feature hierarchies for accurate object detection and semantic
  segmentation.
\newblock In {\em Proceedings of CVPR}, 2014.

\bibitem{dense_net_dpm}
R.~B. Girshick, F.~N. Iandola, T.~Darrell, and J.~Malik.
\newblock Deformable part models are convolutional neural networks.
\newblock {\em CoRR}, 2014.

\bibitem{openMPCFF}
P.~E. Hadjidoukas, V.~V. Dimakopoulos, M.~Delakis, and C.~Garcia.
\newblock A high-performance face detection system using openmp.
\newblock {\em Concurrency and Computation: Practice and Experience}, 2009.

\bibitem{vector_boost}
C.~Huang, H.~Ai, Y.~Li, and S.~Lao.
\newblock Vector boosting for rotation invariant multi-view face detection.
\newblock In {\em Proceedings of ICCV}, 2005.

\bibitem{rotation_invariant}
C.~Huang, H.~Ai, Y.~Li, and S.~Lao.
\newblock High-performance rotation invariant multiview face detection.
\newblock {\em IEEE Transactions on Pattern Analysis and Machine Intelligence},
  2007.

\bibitem{densenet}
F.~Iandola, M.~Moskewicz, S.~Karayev, R.~Girshick, T.~Darrell, and K.~Keutzer.
\newblock Densenet: Implementing efficient convnet descriptor pyramids.
\newblock {\em arXiv preprint arXiv:1404.1869}, 2014.

\bibitem{fddb}
V.~Jain and E.~Learned-Miller.
\newblock Fddb: A benchmark for face detection in unconstrained settings.
\newblock Technical Report UM-CS-2010-009, University of Massachusetts,
  Amherst, 2010.

\bibitem{caffe}
Y.~Jia, E.~Shelhamer, J.~Donahue, S.~Karayev, J.~Long, R.~Girshick,
  S.~Guadarrama, and T.~Darrell.
\newblock Caffe: Convolutional architecture for fast feature embedding.
\newblock {\em arXiv preprint arXiv:1408.5093}, 2014.

\bibitem{Tompson14}
Y.~L. Jonathan J.~Tompson, Arjun~Jain and C.~Bregler.
\newblock Joint training of a convolutional network and a graphical model for
  human pose estimation.
\newblock In {\em Proceedings of NIPS}, 2014.

\bibitem{spp}
S.~R. Kaiming~He, Xiangyu~Zhang and J.~Sun.
\newblock Spatial pyramid pooling in deep convolutional networks for visual
  recognition.
\newblock In {\em Proceedings of ECCV}, 2014.

\bibitem{alex-net}
A.~Krizhevsky, I.~Sutskever, and G.~E. Hinton.
\newblock Imagenet classification with deep convolutional neural networks.
\newblock In {\em Proceedings of NIPS}, 2012.

\bibitem{spatial_pyramid_pooling}
S.~Lazebnik, C.~Schmid, and J.~Ponce.
\newblock Beyond bags of features: Spatial pyramid matching for recognizing
  natural scene categories.
\newblock In {\em Proceedings of CVPR}, 2006.

\bibitem{AFLW}
P.~M.~R. Martin~Koestinger, Paul~Wohlhart and H.~Bischof.
\newblock Annotated facial landmarks in the wild: A large-scale, real-world
  database for facial landmark localization.
\newblock In {\em Proceedings of IEEE International Workshop on Benchmarking
  Facial Image Analysis Technologies}, 2011.

\bibitem{head_hunter}
M.~Mathias, R.~Benenson, M.~Pedersoli, and L.~{Van Gool}.
\newblock Face detection without bells and whistles.
\newblock In {\em Proceedings of ECCV}, 2014.

\bibitem{Osadchy07}
M.~Osadchy, Y.~L. Cun, and M.~L. Miller.
\newblock Synergistic face detection and pose estimation with energy-based
  model.
\newblock In {\em Proceedings of NIPS}, 2005.

\bibitem{Osadchy04}
R.~Osadchy, M.~Miller, and Y.~LeCun.
\newblock Synergistic face detection and pose estimation with energy-based
  model.
\newblock In {\em Proceedings of NIPS}, 2004.

\bibitem{tsm}
D.~Ramanan.
\newblock Face detection, pose estimation, and landmark localization in the
  wild.
\newblock In {\em Proceedings of CVPR}, 2012.

\bibitem{embeddedCFF}
S.~Roux, F.~Mamalet, and C.~Garcia.
\newblock Embedded convolutional face finder.
\newblock In {\em Proceedings of IEEE International Conference on Multimedia
  and Expo}, 2006.

\bibitem{Rowley98}
H.~Rowley, S.~Baluja, and T.~Kanade.
\newblock Neural network-based face detection.
\newblock In {\em Proceedings of CVPR}, 1996.

\bibitem{multires_cascade}
M.~Saberian and N.~Vasconcelos.
\newblock Multi-resolution cascades for multiclass object detection.
\newblock In {\em Proceedings of NIPS}. 2014.

\bibitem{overfeat}
P.~Sermanet, D.~Eigen, X.~Zhang, M.~Mathieu, R.~Fergus, and Y.~LeCun.
\newblock Overfeat: Integrated recognition, localization and detection using
  convolutional networks.
\newblock In {\em Proceedings of International Conference on Learning
  Representations}, 2014.

\bibitem{deepid2}
Y.~Sun, Y.~Chen, X.~Wang, and X.~Tang.
\newblock Deep learning face representation by joint
  identification-verification.
\newblock In {\em Proceedings of NIPS}. 2014.

\bibitem{googlenet}
C.~Szegedy, W.~Liu, Y.~Jia, P.~Sermanet, S.~Reed, D.~Anguelov, D.~Erhan,
  V.~Vanhoucke, and A.~Rabinovich.
\newblock Going deeper with convolutions.
\newblock {\em CoRR}, 2014.

\bibitem{google_detector}
C.~Szegedy, S.~Reed, D.~Erhan, and D.~Anguelov.
\newblock Scalable, high-quality object detection.
\newblock {\em CoRR}, 2014.

\bibitem{google_2014}
C.~Szegedy, S.~Reed, D.~Erhan, and D.~Anguelov.
\newblock Scalable, high-quality object detection.
\newblock {\em CoRR}, 2014.

\bibitem{deepface}
Y.~Taigman, M.~Yang, M.~Ranzato, and L.~Wolf.
\newblock Deepface: Closing the gap to human-level performance in face
  verification.
\newblock In {\em Proceedings of CVPR}, 2014.

\bibitem{torralba_cascade}
A.~Torralba, K.~Murphy, and W.~Freeman.
\newblock Sharing visual features for multiclass and multiview object
  detection.
\newblock {\em IEEE Transactions on Pattern Analysis and Machine Intelligence},
  2007.

\bibitem{selective_search}
J.~Uijlings, K.~van~de Sande, T.~Gevers, and A.~Smeulders.
\newblock Selective search for object recognition.
\newblock {\em International Journal of Computer Vision}, 2013.

\bibitem{Vaillant93}
R.~Vaillant, C.~Monrocq, and Y.~LeCun.
\newblock An original approach for the localisation of objects in images.
\newblock In {\em Proceedings of International Conference on Artificial Neural
  Networks}, 1993.

\bibitem{Vaillant94}
R.~Vaillant, C.~Monrocq, and Y.~LeCun.
\newblock Original approach for the localisation of objects in images.
\newblock {\em IEE Proc on Vision, Image, and Signal Processing}, 1994.

\bibitem{VJ_multi_view}
M.~Viola and P.~Viola.
\newblock Fast multi-view face detection.
\newblock In {\em Proceedings of CVPR}, 2003.

\bibitem{VJ}
P.~Viola and M.~J. Jones.
\newblock Robust real-time face detection.
\newblock {\em International Journal of Computer Vision}, 2004.

\bibitem{parallel_cascade}
B.~Wu, H.~Ai, C.~Huang, and S.~Lao.
\newblock Fast rotation invariant multi-view face detection based on real
  adaboost.
\newblock In {\em IEEE International Conference on Automatic Face and Gesture
  Recognition}, 2004.

\bibitem{structural_model}
J.~Yan, X.~Zhang, Z.~Lei, and S.~Li.
\newblock Face detection by structural models.

\end{thebibliography}

% sigproc.bib is the name of the Bibliography in this case
% ACM needs 'a single self-contained file'!
%\balancecolumns

%\balancecolumns % GM June 2007
% That's all folks!
\end{document}